\documentclass[10pt,twocolumn,letterpaper]{article}

\usepackage{iccv}              
\usepackage[accsupp]{axessibility}  

\definecolor{iccvblue}{rgb}{0.21,0.49,0.74}
\usepackage[pagebackref,breaklinks,colorlinks,allcolors=iccvblue]{hyperref}
\usepackage{booktabs}
\usepackage{multirow}
\usepackage{graphicx}
\usepackage{bm}
\usepackage{float}

\def\paperID{APAH-8} 
\def\confName{ICCV}
\def\confYear{2025}

\title{Automated C-Arm Positioning via Conformal Landmark Localization}

\author{Ahmad Arrabi$^{1}$ \hspace{0.15cm} Jay Hwasung Jung$^{1}$ \hspace{0.15cm} Jax Luo$^{2}$ \hspace{0.15cm} Nathan Franssen$^{3}$ \hspace{0.15cm} Scott Raymond$^{2}$ \hspace{0.15cm} Safwan Wshah$^{1}$
\and
{University of Vermont, Department of Computer Science, Burlington VT, USA$^{1}$}
\and
{Cleveland Clinic, Neurological Institute, Cleveland OH, USA$^{2}$}
\and
{University of Vermont Medical Center, Burlington VT, USA$^{3}$}
\and
\tt\footnotesize\{aarrabi, jjung2, swshah\}@uvm.edu, \{luoj2, raymons3\}@ccf.org, \{nathan.franssen\}@uvmhealth.org
}

\begin{document}
\maketitle
\begin{abstract}
Accurate and reliable C-arm positioning is essential for fluoroscopy-guided interventions. However, clinical workflows rely on manual alignment that increases radiation exposure and procedural delays. In this work, we present a pipeline that autonomously navigates the C-arm to predefined anatomical landmarks utilizing X-ray images. Given an input X-ray image from an arbitrary starting location on the operating table, the model predicts a 3D displacement vector toward each target landmark along the body. To ensure reliable deployment, we capture both aleatoric and epistemic uncertainties in the model's predictions and further calibrate them using conformal prediction. The derived prediction regions are interpreted as 3D confidence regions around the predicted landmark locations. The training framework combines a probabilistic loss with skeletal pose regularization to encourage anatomically plausible outputs. We validate our approach on a synthetic X-ray dataset generated from DeepDRR. Results show not only strong localization accuracy across multiple architectures but also well-calibrated prediction bounds. These findings highlight the pipeline's potential as a component in safe and reliable autonomous C-arm systems. Code is available at \url{https://github.com/AhmadArrabi/C_arm_guidance_APAH}
\vspace{-5mm}
\end{abstract}

\section{Introduction}
\label{sec:intro}
C-arm machines are used in interventional procedures that require fluoroscopy. These machines allow real-time radiographic projections during operations such as vascular access and orthopedic repairs. Before beginning a procedure, clinicians manually position the C-arm over a region of interest (ROI) on the patient~\cite{shao2014virtual}. This manual positioning step is often guided by repeated fluoroscopy exposures, which can introduce delays and unnecessary radiation for both patients and clinicians. Moreover, the need for quick and precise positioning becomes critical in urgent procedures, e.g., stroke and trauma, which may be performed by less experienced personnel in low-resource settings~\cite{stein2021correlations}. These challenges highlight the need for assistive technologies that enable safe, efficient, and reliable fluoroscopy setup and operation.
 
Automated fluoroscopy positioning can help imaging workflows by utilizing predefined anatomical ROIs that are frequently targeted in procedures. For example, during a stroke thrombectomy, the ROI involves cerebral vessels, such as the internal (ICA) and middle cerebral artery (MCA)~\cite{potter2019ct}. In orthopedic trauma surgeries, the ROI would be around the area of injury, e.g., specific bones or joints~\cite{clevelandclinic2022orthopedic}. Aligning the C-arm machine with the target ROI can be formalized as a localization task, where the system needs to maneuver the C-arm to a given anatomical landmark~\cite{KashkoushA193}.

Introducing any system of this kind requires providing both accuracy and a confidence measure. However, deep learning methods often produce deterministic point estimates that lack any measure of uncertainty. To address this gap, we propose a pipeline designed for automatic C-arm positioning. Our method models both the aleatoric uncertainty (inherent variability in the data) and epistemic uncertainty (model's lack of knowledge)~\cite{uncertainty_1,uncertainty_2}. Aleatoric uncertainty is learned by training the model to output Gaussian-distributed predictions and minimizing the negative log-likelihood loss. Epistemic uncertainty is estimated using Monte Carlo Dropout (MCD), by averaging predictions from multiple stochastic forward passes.

Additionally, we integrate conformal prediction, a distribution-free, post-hoc method that constructs prediction intervals with formal coverage guarantees under the assumption of data exchangeability. In our context, these intervals take the form of 3D spheres around predicted landmarks, defining a region where the true location is expected to lie with a predefined probability ($1-\alpha$ in~\cref{sec:methods}). The radius of each sphere adapts dynamically based on the total predicted uncertainty, enabling the system to express per-sample confidence. This offers a step toward clinically viable, risk-aware systems in fluoroscopy.

We evaluate our method on a synthetic X-ray dataset generated using DeepDRR~\cite{unberath2018deepdrrcatalystmachine} from CT scans collected from the New Mexico Decedents Image Database (NMDID)~\cite{edgar2020nmdid}. Our results demonstrate not only strong localization performance but also accurate and calibrated uncertainty estimates. We also present a stroke thrombectomy use case, where the model navigates to the skull using different paths. The main contributions of this work can be summarized as follows:
\begin{itemize}
\item We introduce a general uncertainty-aware pipeline for automated C-arm positioning, specifically to predict anatomical landmarks with quantified confidence.
\item Utilizing conformal prediction, our method provides formal statistical guarantees, ensuring that true landmark locations lie within calibrated bounds.
\item We validate our approach through extensive experiments, including a practical evaluation on a simulated stroke thrombectomy use case.
\end{itemize}
\section{Related Work}
\label{sec:related_work} 
Computer vision has been applied in medical imaging for tasks such as organ segmentation, lesion detection, and image registration~\cite{ZOU2023100003, loftus2022uncertainty, doi:10.1148/radiol.222217}. While these approaches have shown strong performance in diagnostic tasks, they are often unsuitable for time-sensitive clinical applications due to their limited interpretability and lack of uncertainty understanding~\cite{kompa2021second, ZOU2023100003, loftus2022uncertainty, doi:10.1148/radiol.222217}. 

\noindent\textbf{Image-Guided C-arm Automation:} 
Image-guided C-arm control has been explored in interventional procedures, where accurate positioning to ROIs is critical. To overcome the limitations of traditional approaches, such as manual or VR- and AR-based control~\cite{shao2014virtual,unberath2018augmented}, deep learning-based methods have emerged.

~\cite{Kausch2020} proposed a model for automatic C-arm positioning, targeting anatomy-specific views in orthopedic surgery. Their approach uses a two-stage network to predict 5 Degrees of Freedom (DoF) pose updates from a single synthetic DRR image. They focus on localizing the proximal femur (PF) and the fourth lumbar vertebra (LV4), demonstrating robust performance across anatomical variability using synthetic training data. However, its limited anatomical ROI and reliance on multi-stage predictions limit its broader clinical applicability. Another study in~\cite{esfandiari2020deep} introduced C-arm pose estimation from a single view. DeepDRR~\cite{unberath2018deepdrrcatalystmachine} was used to produce realistic X-ray images of the pelvis. Their approach achieved competitive results, but the use of a limited dataset raises concerns about the model's generalizability, not only to other ROIs but also across diverse patient anatomies.

To move beyond region-specific models,~\cite{10980945} proposed a self-supervised approach for C-arm landmark classification and regression. Their method uses a pretext regression task to predict the location of input X-ray images across the entire body and a downstream landmark classification task. However, no navigation or control was implemented. These simplifications raise questions about the method's reliability when applied to real-world settings.

\noindent\textbf{Uncertainty Quantification in Medical Vision:} 
Artificial intelligence (AI) has become a powerful assistive tool for decision-making across various domains~\cite{10879299, jiang2018trusttrustclassifier}. Despite its efficiency, the reliability of AI predictions remains questionable due to the underlying uncertainty derived from data variability, model miscalibration~\cite{Abdar_2021}. As a result, many researchers argue that trust in AI systems should be supported by proper uncertainty quantification~\cite{jiang2018trusttrustclassifier, huang2023quantifyingepistemicuncertaintydeep, smith2024rethinkingaleatoricepistemicuncertainty,uncertainty_3,uncertainty_4}. We summarize the most common uncertainty quantification approaches below~\cite{Abdar_2021}: 
\begin{itemize}[leftmargin=*]
    \item \textbf{Bayesian inference}: These approaches quantify uncertainty by treating the network weights as probability distributions rather than point estimates. They capture epistemic uncertainty by defining approximations over the weight posterior integral. These methods include variational inference or Markov Chain Monte Carlo methods~\cite{Abdar_2021, DBLP:journals/corr/abs-2006-15172, kumar2024uncertaintyawaredeepneuralrepresentations}.
    \item \textbf{Ensemble techniques}: Ensembles estimate uncertainty by training multiple models independently and combining their predictions. The diversity among the models reflects epistemic uncertainty~\cite{Abdar_2021, kumar2024uncertaintyawaredeepneuralrepresentations}.
    \item \textbf{Monte Carlo dropout}: Monte Carlo dropout estimates uncertainty by applying dropout during both training and testing~\cite{gal2016dropoutbayesianapproximationrepresenting}. Dropout randomly deactivates parts of the network, creating slightly different models each time. By running multiple forward passes with different dropout masks at test time, the method generates varying predictions~\cite{Abdar_2021,kumar2024uncertaintyawaredeepneuralrepresentations}.
\end{itemize}
In medical imaging tasks such as segmentation and registration, uncertainty quantification provides confidence estimates that can be used as additional information in the decision-making process. This has motivated the integration of uncertainty estimation into deep learning models across various imaging modalities, such as magnetic resonance imaging (MRI), computed tomography (CT), and X-ray~\cite{ZOU2023100003, loftus2022uncertainty, doi:10.1148/radiol.222217}. 
The integration of uncertainty quantification into C-arm automation remains largely unexplored, highlighting the novelty of our approach.
\section{Dataset}
\label{sec:dataset}
Training data for C-arm positioning is limited due to the practical and ethical constraints of obtaining large-scale fluoroscopic images from real patients. As a result, existing datasets \cite{Wang2017ChestXray8, Gonzalez2020SIMBA} are often focused on specific anatomical regions, e.g., chest, hands, pelvis, making it challenging to train deep learning models that continuously generalize across the entire body. Thus, we constructed a synthetic X-ray dataset covering the head, neck, and upper extremities from computed tomography (CT) data using Digitally Reconstructed Radiographs (DRRs). The collected data is summarized in~\cref{tab:dataset_summary}.
\begin{table}[t]
\centering
\caption{Dataset summary. The dataset was randomly split into 70\% for training, 15\% for calibration, and 15\% for testing, on the patient level.}
\label{tab:dataset_summary}
\resizebox{\columnwidth}{!}{%
\begin{tabular}{l l}
\toprule
\textbf{Property} & \textbf{Value} \\
\midrule
\# CT scans & 260 \\
Samples per CT volume & 1024 X-ray images\\
Total samples & 266{,}240 X-ray images \\
\# landmarks (ROIs) & 14 \\
Vertical sampling & Uniform \\
Horizontal sampling & Gaussian ($\sigma = 47.5\text{mm}$) \\
Depth sampling & Gaussian ($\sigma = 100\text{mm}$) \\
\bottomrule
\end{tabular}}
\vspace{-5mm}
\end{table}

\noindent\textbf{CT data Acquisition: }
To construct a dataset that enables learning across the continuous anatomical structure of the human body, we utilize CT scans from the publicly available New Mexico Decedent Image Database (NMDID)~\cite{edgar2020nmdid}. For this study, we selected 260 CT scans to generate synthetic DRR images. To ensure comprehensive anatomical coverage, we use scans covering the head, neck, and upper extremities, captured with 3 mm slice thickness and 3 mm spacing. Each slice was stored in DICOM (.dcm) format, but for processing efficiency, we converted these slices into one unified file with NIfTI (.nii) format. 

To ensure consistent spatial representation, all CT scans were normalized by mapping their physical coordinates into the unit cube \([0,1]^3\) using each scan’s own dimensions. This transformation preserves the relative spatial relationships between anatomical structures while allowing the model to learn from scale-invariant spatial patterns.

\noindent\textbf{DRR Image Generation: }
A DRR is a synthetic 2D projection of a 3D CT volume. To ensure the realism and clinical applicability of our training data, we use DeepDRR~\cite{unberath2018deepdrrcatalystmachine} - a deep learning-based framework designed to accurately simulate the C-arm machine. It takes a 3D CT Volume and isocenter (a fixed reference point in 3D space that defines the center of the projection) as inputs, and outputs a synthetic X-ray image.
\begin{figure}[!htbp]
    \centering
    \includegraphics[width=0.9\linewidth]{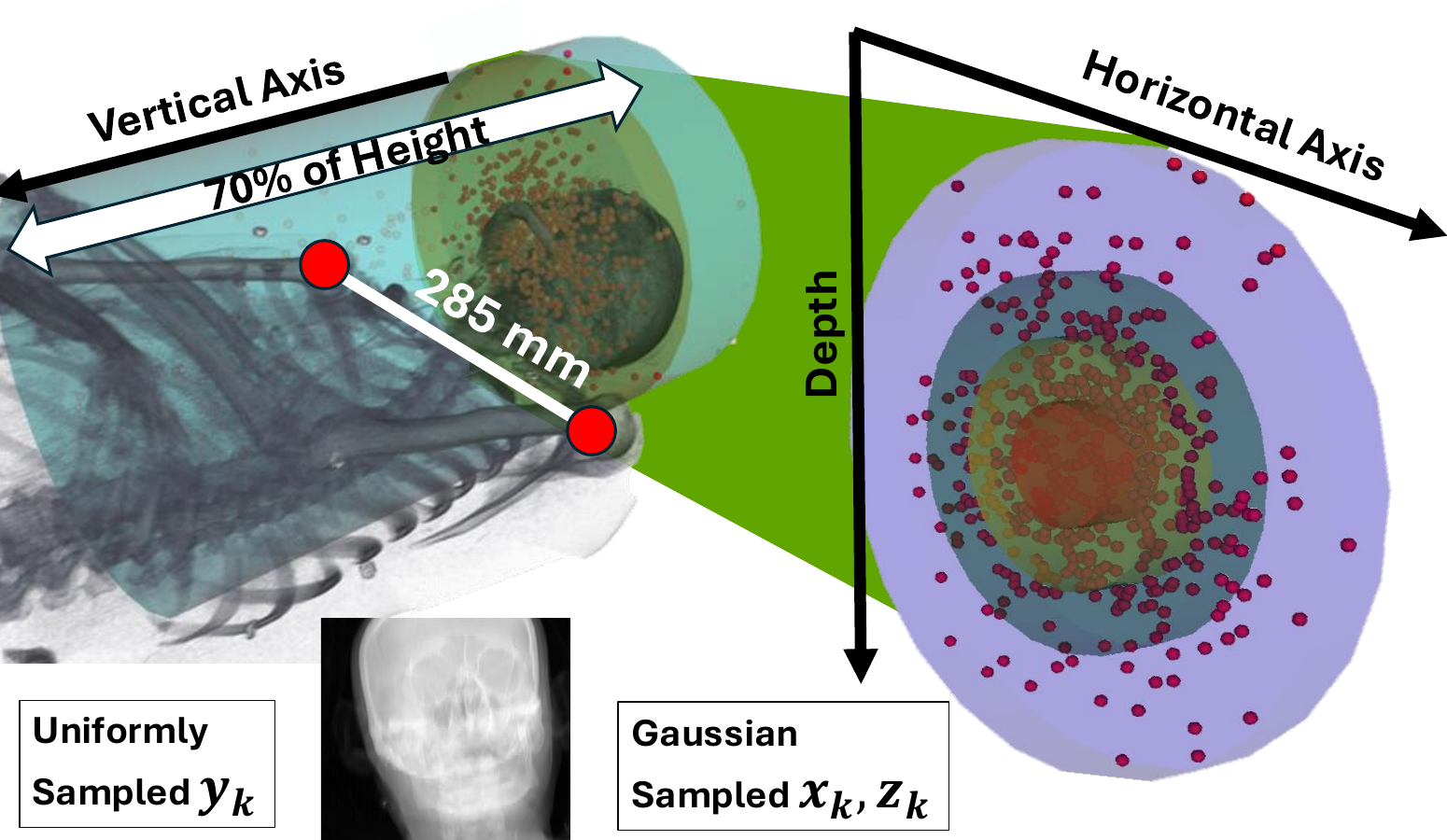}
    \caption{Overview of the sampling process. We densely sample each CT volume by defining independent distributions across each spatial axis.}
    \label{fig:sampling}
\end{figure}

\noindent\textbf{Sampling Process: }
We systematically generate synthetic X-ray images from multiple viewpoints using DeepDRR. Each view is simulated by positioning the C-arm system around a fixed anatomical isocenter, indicated by the red dot in~\cref{fig:sampling}.

To introduce spatial variability and create a rich training dataset,  we densely sampled isocenters within the CT volume to accurately simulate the practical workflow of a typical C-arm. Specifically, the vertical (superior-inferior) position is sampled uniformly within 70\% of the anatomical height. The horizontal (left-right) position is sampled with a Gaussian spread centered around the anatomy, with a standard deviation chosen to reflect the average width (285 mm) of the annotated left and right humeral heads. The depth (anterior-posterior) also follows a Gaussian distribution, with a broader spread ($\sigma = 100\text{ mm}$).

These sampling strategies not only capture natural variations in views but also ensure that projections avoid empty regions and focus on the torso where the ROIs are most likely to be found. Each CT scan was sampled 1,024 times, resulting in a total of 266,240 synthetic images.
\begin{figure}[!b]
    \centering
    \includegraphics[width=\linewidth]{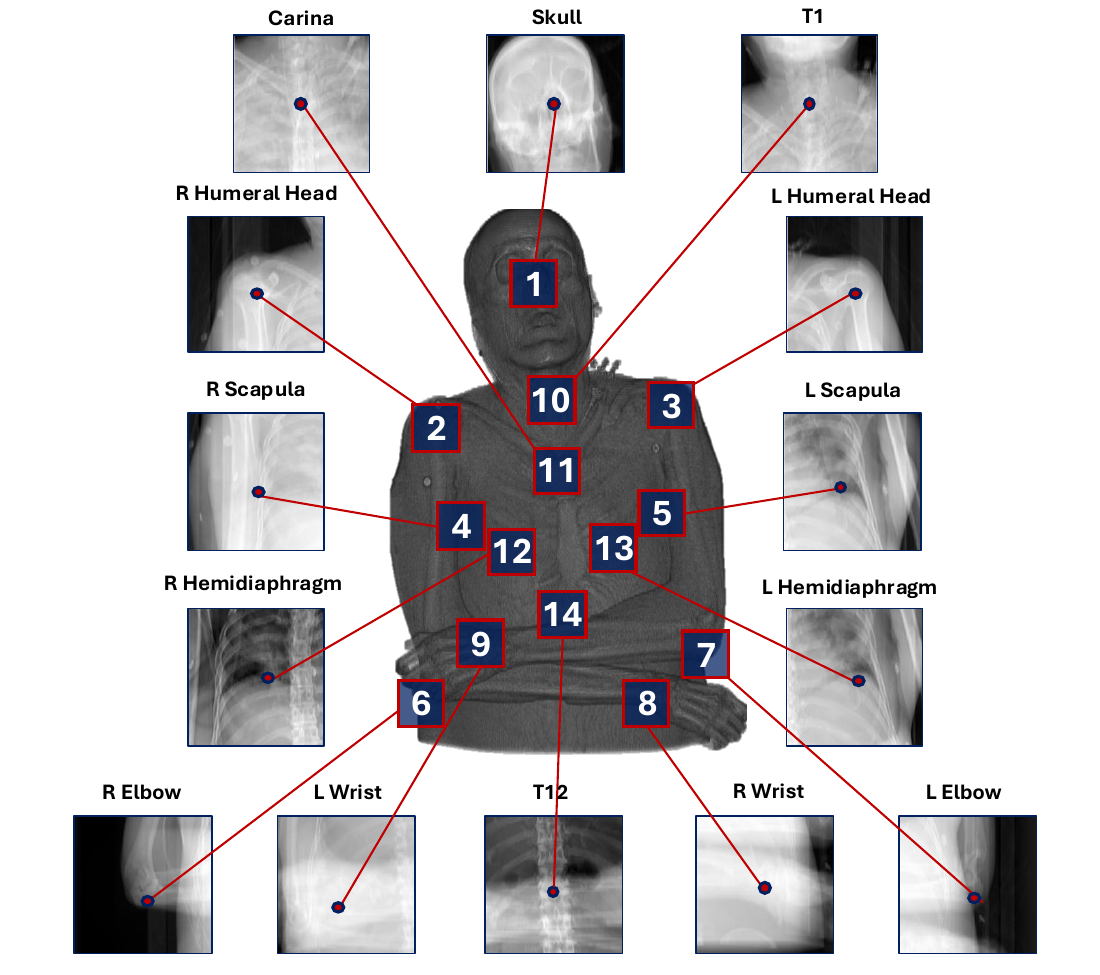}
    \caption{Visualization of landmark annotations. These landmarks represent ROI used across different clinical applications.}
    \label{fig:landmark}
\end{figure}

\noindent\textbf{Landmark Annotations: }
To create ROI annotations, we define 14 landmarks that span the 3D volume and provide substantial anatomical coverage, shown in~\cref{fig:landmark}. This enables us to assess per-landmark uncertainty when guiding C-arm machines from arbitrary initial positions to predefined ROIs. To facilitate landmark annotations from each CT scan, we utilized the annotation tool introduced in~\cite{10980945}. In total, we collected 3,643 landmark coordinates from 260 patient CT scans.

\noindent\textbf{Dataset split: }
We partition the dataset into 70\% for training and 15\% for each calibration and testing. This corresponds to 182 CT volumes used in training, and 39 each for calibration and testing. The need for a calibration set comes from our application of split conformal prediction, which requires held-out data to construct prediction intervals with formal statistical coverage guarantees in testing. 
\section{Methodology}
\label{sec:methods}
\begin{figure*}[!t]
    \centering
    \includegraphics[width=0.9\linewidth]{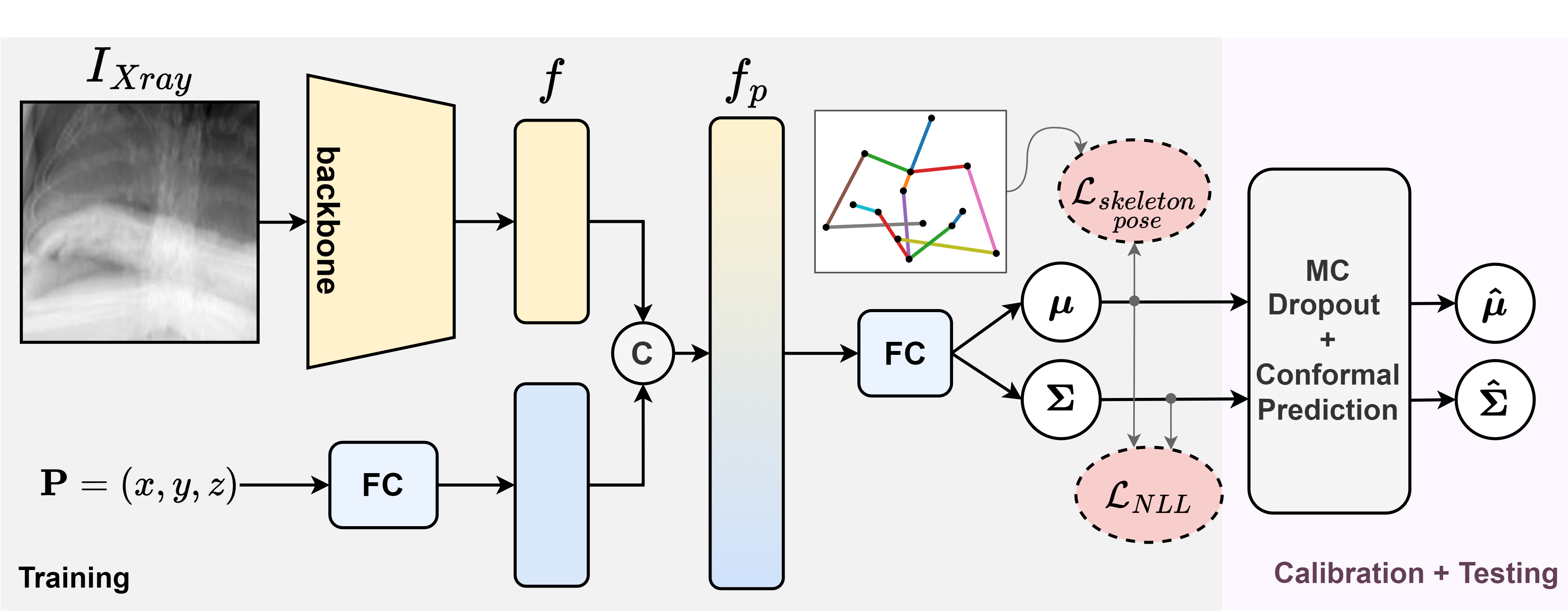}\caption{Overview of the proposed pipeline. The input X-ray is encoded using a backbone network, and the C-arm position is embedded into the image features. For each of the 14 anatomical landmarks, the model outputs the parameters of a 3D Gaussian: a mean vector $\boldsymbol{\mu} = (\mu_{\Delta x}, \mu_{\Delta y}, \mu_{\Delta z})$ and a diagonal covariance matrix $\boldsymbol{\Sigma} = \text{diag}(\sigma^2_{\Delta x},\sigma^2_{\Delta y},\sigma^2_{\Delta z})$, leading to $14\times3\times2$ output neurons. A skeleton pose loss further regularizes predictions based on the patient's prior anatomical topology. In Calibration and testing, we apply Monte Carlo dropout with $T$ stochastic forward passes to estimate the total uncertainty.}
    \label{fig:pipeline}
    \vspace{-5mm}
\end{figure*}
We formalize our problem as a regression task aimed at predicting C-arm control. Given an input X-ray image $I_{Xray} \in \mathbb{R}^{3\times h\times w}$ (where $h$ and $w$ denote height and width, respectively) captured from an arbitrary location $\bm{P}=(x,y,z)\in[0,1]^3$ within a normalized 3D space. We predict the needed displacements in all positional axes to reach a set of predefined target landmarks $\{\Delta \hat{\bm{P}_k}=(\Delta \hat{x}_k,\Delta \hat{y}_k,\Delta \hat{z}_k) \ :k \in \{1,\dots , m \} \}$, where $m$ is the number of landmarks. These landmarks represent ROIs where operators typically maneuver the C-arm (see~\cref{sec:dataset}). \textit{Note that for readability, we will consider a single landmark in our notation, but our pipeline processes multiple independent landmark displacements.}

~\cref{fig:pipeline} illustrates the proposed pipeline. The input image is first projected into a latent representation $f\in \mathbb{R}^{c\times h_f \times w_f}$ by a backbone network. This representation is then passed to a positioning encoding module that concatenates it with a positional embedding vector that encodes the C-arm pose $f_p =cat(f, \ linear(\bm{P}))$. Finally, the regression head (sequence of linear layers) outputs the predicted displacements $\{\Delta \hat{\bm{P}_k}\}$. Note that in all linear layers, SiLU activation was used, followed by a dropout layer.

\subsection{Uncertainty Modeling}
In deep learning literature, it is widely accepted that the total predictive uncertainty can be approximated as the sum of aleatoric and epistemic uncertainties~\cite{uncertainty_1,uncertainty_2}. These two arise from fundamentally different sources, so following prior work, we assume independence between the two, and define the total uncertainty as follows,
\begin{equation}
    \label{eq:uncertainity}
    \sigma_{total}^2 = \sigma_{epestimic}^2 + \sigma_{aleatoric}^2
\end{equation}

To estimate $\sigma^2_{aleatoric}$, we model the regression outputs as independent Gaussian distributions, replacing point estimates with predictive distributions. For each landmark, the network predicts the parameters for its mean vector $\boldsymbol{\mu_k} =(\mu_{\Delta x},\mu_{\Delta y},\mu_{\Delta z})$ and its diagonal covariance matrix $\boldsymbol{\Sigma_k} = \text{diag}(\sigma^2_{\Delta x},\sigma^2_{\Delta y},\sigma^2_{\Delta z})$, so the predicted displacement is modeled as in~\cref{eq:target}. This formulation captures the inherent variance of the data, e.g., anatomical variability and landmark annotation inconsistencies. 
\begin{equation}
    \label{eq:target}
    \Delta \hat{\mathbf{P}}_k \sim \mathcal{N}(\boldsymbol{\mu}_k, \mathbf{\Sigma}_k)
\end{equation}

To have a comprehensive understanding of uncertainty, we account for $\sigma^2_{epestimic}$, which reflects the uncertainty over the model's parameters. We adopt Monte Carlo Dropout (MCD), a lightweight Bayesian approximation that interprets dropout as a form of variational inference~\cite{dropout}. In MCD, dropout layers are activated in both training and testing, which transforms any deterministic network into a probabilistic one, where each forward pass samples a different subnetwork by randomly deactivating weights with probability $p$.

In~\cref{fig:pipeline}, dropout layers are added after all linear layers, making our pipeline stochastic. Every forward pass yields a different output due to the random masking of weights. This variability across multiple passes for the same input is used to estimate the variance of the model's parameters. Therefore, we run $T$ forward passes for each input. The final prediction is the mean value of those runs, while their variance is $\sigma^2_{epestimic}$, which is added with the predicted variance as in~\cref{eq:uncertainity}.  

\subsection{Conformal Prediction}
Conformal prediction is a post-hoc uncertainty quantification framework that produces prediction sets with guaranteed statistical coverage, with no assumptions about the underlying model~\cite{conformal_1}. We utilize a split conformal prediction method where we use 15\% of the dataset to calibrate the nonconformity scores~\cite{uncertainty_2}. We use the Euclidean error, defined as the distance between predicted and ground-truth landmark positions, to construct prediction regions. Specifically, we compute the empirical distribution of these errors over the calibration set and extract quantiles corresponding to the desired $1-\alpha$ confidence levels. These quantiles define the radii of fixed spherical regions around each prediction. In our proposed pipeline, conformal prediction complements the model's uncertainty estimates, as it produces statistically valid confidence regions around predictions.

\noindent\textbf{Nonconformity Score:} To adapt the prediction region to the predicted uncertainty of each test sample, inspired by~\cite{conformal_3}, we develop a dynamic nonconformity score based on the model's predicted uncertainty, $\sigma^2_{total}$ in~\cref{eq:uncertainity}. Since our task involves 3D landmark localization, we define the nonconformity score using a normalized Euclidean error between the predictions and ground-truth landmark displacements. For a sample $i$ and landmark $k$, the nonconformity score is given by,
\begin{equation}
    \label{nonconformity}
    s^{(i)}_k = \frac{d(\Delta \hat{\bm{P}}_k^{(i)},  \ \Delta\bm{P}_k^{(i)})}{\sigma_{k \ total}^{2 \ (i)}}
\end{equation}
Where $\Delta \hat{\bm{P}}_k^{(i)}$ denotes the predicted 3D displacement to reach landmark $k$, $\Delta\bm{P}_k^{(i)}$ is the ground truth displacement, $\sigma_{k \ total}^{2 \ (i)}$ is the total predicted uncertainty of the model, $d(.,.)$ is the Euclidean distance. After computing these scores for all samples in the calibration set, we extract the empirical quantiles $Q^k_{1-\alpha}$ for predefined $\alpha$ values.

\noindent\textbf{Prediction Region:} At test time, the prediction region for each landmark $k$ is defined using the model's predicted uncertainty and the precomputed quantile $Q^k_{1-\alpha}$, as follows,
\begin{equation}
    \label{eq:region}
    R_k^{i} = \{\bm{P} \in \mathbb{R}^3: d(\bm{P}, \Delta \hat{\bm{P}}_k^{(i)}) \le \sigma_{k \ total}^{2 \ (i)} \cdot Q^k_{1-\alpha}\}
\end{equation}
This region corresponds to a 3D sphere centered at the predicted displacement $\Delta \hat{\bm{P}}_k^{(i)}$, whose radius scales with the model's uncertainty. Ensuring that the true landmark location lies within this region with probability $1-\alpha$, which provides interpretable and risk-aware predictions.

\subsection{Training Objectives}
\noindent\textbf{Negative Log-Likelihood Loss:} As each regression target is modeled as a Gaussian, it is possible to perform Maximum Likelihood Estimation (MLE) through minimizing the negative log-likelihood (NLL) loss. Note that with our assumption of independence between landmarks and positional axes, the predicted covariance matrix is diagonal, which simplifies the NLL derivation. For a given landmark $k$, the NLL loss is given by,
\begin{equation}
    \label{eq:nll}
    \begin{split}
        \mathcal{L}_{NLL} &= -\text{log } p(\Delta \hat{\bm{P}}| \boldsymbol{\mu}, \boldsymbol{\Sigma}) \\
        &= \frac{1}{2}\beta \ \text{log}\boldsymbol{|\Sigma|} + \frac{1}{2}(\Delta \hat{\bm{P}}-\boldsymbol{\mu})^T\boldsymbol{|\Sigma|}(\Delta \hat{\bm{P}}-\boldsymbol{\mu}) + \gamma \\
        &= \frac{1}{2}\sum_{i\in\{x,y,z\}}\Big(\beta \text{ log } \sigma_i^2 + \frac{(\Delta \hat{P}_i - \mu_{\Delta i})^2}{\sigma_i^2}\Big) + \gamma
    \end{split}
\end{equation}
Where $\Delta \hat{\bm{P}} = (\mu_{\Delta x}, \mu_{\Delta y}, \mu_{\Delta z})$, $\beta$ is a constant that regularizes the variance prediction, and $\gamma$ is a constant term from the Gaussian likelihood function which can be omitted during optimization.

\noindent\textbf{Skeleton Pose Loss:}
Landmarks are spatially correlated and constrained by skeletal geometry. To incorporate this structural prior, we introduce a skeleton pose loss that regularizes the model’s predictions by enforcing consistency with a predefined skeletal topology. For each CT scan, a custom skeletal graph is constructed from its landmark annotations. An example topology is illustrated in~\cref{fig:pipeline}, capturing intuitive dependencies among landmarks in a humanoid form (e.g., wrists connected to shoulders, and the Carina, T1, and skull forming a central axis). We represent the skeleton pose as an undirected graph, where each node is a landmark, and edges represent connections. The loss is defined as the distance between the predicted graph and the ground truth skeleton graph. The loss is formally introduced as follows,
\begin{equation}
    \mathcal{L}_{skeleton \ pose} = \sum_{(i,j)\in G} \Big( d(\hat{\bm{P}_i}, \hat{\bm{P}}_j) - d(\bm{P}_i, \bm{P}_j) \Big)
\end{equation}
Where $G$ is the prior skeleton graph, $\hat{\bm{P}}$ is the location of the C-arm after the displacement, and $\bm{P}$ is the actual landmark location. The total training loss used in our experiments is given by,
\begin{equation}
    \label{eq:loss}
    \mathcal{L} = \mathcal{L}_{NLL} + \lambda \ \mathcal{L}_{skeleton \ pose}
\end{equation}
Where $\lambda$ is a hyperparameter that determines how much influence the skeleton loss has on training. 

\subsection{Data Augmentation}
In clinical practice, patients' initial positioning varies due to time constraints and workflow differences across procedures. To account for this variability and improve the generalizability of the model, we incorporate a patient-position augmentation method. This method simulates realistic shifts in patient placement by randomly perturbing their position within a controlled range. The augmentation is applied with probability $p_o$ to the patient's $(x,y)$ coordinates as follows,
\begin{equation}
    (x, y) =
    \begin{cases} 
       (x,y) + shift & p\leq p_\circ \\
       (x,y) & p>p_\circ,
   \end{cases}
\end{equation}
Where $shift \sim \mathcal{U}(-\eta, \ \eta)$ as $\eta$ controls the augmentation strength: weak, medium, or strong.
\section{Experiments}
\label{sec:experiments}
\begin{table}[t]
\centering
\caption{Pipeline performance with different backbones. In all backbones, the given calibration intervals are reliable, as PRCP values are close to their $1-\alpha$ levels. }
\label{tab:backbones}
\resizebox{\columnwidth}{!}{%
\begin{tabular}{@{}lcccccc@{}}
\toprule
\multicolumn{1}{c}{\multirow{2}{*}{Backbones}} &
  \multirow{2}{*}{\begin{tabular}[l]{@{}c@{}}Mean distance \\ from GT (mm) $\downarrow$ \end{tabular}} &
  \multicolumn{2}{c}{NLL $\downarrow$} &
  \multicolumn{3}{c}{PRCP} \\ \cmidrule(lr){3-4}\cmidrule(lr){5-7}
\multicolumn{1}{c}{} &                & Calibration    & Test           & 90\%    & 95\%    & 97\%   \\ \midrule
ConvNeXt Base        & \textbf{38.60} & \textbf{-2.18} & \textbf{-2.27} & 90.18\% & 95.03\% & 96.92\% \\
Resnet101            & 40.91          & \textbf{-2.18} & -2.21          & 89.74\% & 94.71\% & 96.69\% \\
Resnet34             & 44.04          & -2.11          & -2.13          & 89.18\% & 94.58\% & 96.83\% \\
ViT Base/16          & 42.61          & -2.13          & -2.15          & 89.71\% & 94.58\% & 96.58\% \\ \bottomrule
\end{tabular}}
\vspace{-5mm}
\end{table}
\begin{table*}[t]
\centering
\caption{Performance per landmark. Regions with less movement tend to produce more accurate C-arm positioning, while moving ones are less stable. Most empirical PRCP values match the designed $1-\alpha$ confidence levels.}
\label{tab:landmarks}
\resizebox{2.1\columnwidth}{!}{%
\begin{tabular}{@{}lcccccccccccccccc@{}}
\toprule
\multirow{3}{*}{landmark} & \multicolumn{4}{c}{ConvNeXt Base} & \multicolumn{4}{c}{Resnet101} & \multicolumn{4}{c}{Resnet34} & \multicolumn{4}{c}{ViT Base/16} \\ \cmidrule(lr){2-5} \cmidrule(lr){6-9} \cmidrule(lr){10-13} \cmidrule(lr){14-17} 
 & \multirow{2}{*}{\begin{tabular}[c]{@{}c@{}}Distance from \\GT (mm) $\downarrow$\end{tabular}} & \multicolumn{3}{c}{PRCP} & \multirow{2}{*}{\begin{tabular}[c]{@{}c@{}}Distance from \\GT (mm) $\downarrow$\end{tabular}} & \multicolumn{3}{c}{PRCP} & \multirow{2}{*}{\begin{tabular}[c]{@{}c@{}}Distance from\\GT (mm) $\downarrow$\end{tabular}} & \multicolumn{3}{c}{PRCP} & \multirow{2}{*}{\begin{tabular}[c]{@{}c@{}}Distance from\\GT (mm) $\downarrow$\end{tabular}} & \multicolumn{3}{c}{PRCP} \\ \cmidrule(lr){3-5} \cmidrule(lr){7-9} \cmidrule(lr){11-13} \cmidrule(l){15-17} 
 &  & 90\% & 95\% & 97\% &  & 90\% & 95\% & 97\% &  & 90\% & 95\% & 97\% &  & 90\% & 95\% & 97\% \\ \midrule
1: Skull & \textbf{29.25} & 89.55\% & 94.44\% & 96.55\% & 31.49 & 87.63\% & 93.24\% & 95.62\% & 35.60 & 87.89\% & 93.55\% & 96.07\% & 33.62 & 89.70\% & 94.58\% & 96.39\% \\
2: R Humeral Head & \textbf{29.52} & 88.92\% & 94.46\% & 96.78\% & 32.54 & 87.55\% & 93.34\% & 95.80\% & 35.81 & 86.61\% & 92.70\% & 95.42\% & 34.54 & 88.97\% & 94.29\% & 96.36\% \\
3: L Humeral Head & \textbf{30.04} & 93.73\% & 98.04\% & 99.52\% & 32.99 & 93.75\% & 98.06\% & 99.51\% & 36.42 & 92.49\% & 97.66\% & 99.61\% & 35.31 & 92.31\% & 96.90\% & 98.82\% \\
4: R Scapula & \textbf{38.70} & 86.12\% & 92.79\% & 95.55\% & 40.96 & 85.40\% & 91.55\% & 94.56\% & 43.03 & 84.94\% & 92.23\% & 95.86\% & 42.91 & 86.80\% & 92.78\% & 95.47\% \\
5: L Scapula & \textbf{36.07} & 86.08\% & 92.40\% & 95.00\% & 38.09 & 86.64\% & 92.58\% & 94.99\% & 40.52 & 85.57\% & 92.85\% & 96.05\% & 40.28 & 85.43\% & 91.66\% & 94.62\% \\
6: R Elbow & \textbf{45.94} & 94.61\% & 98.39\% & 99.52\% & 47.53 & 95.35\% & 98.81\% & 99.64\% & 51.80 & 93.88\% & 98.01\% & 99.21\% & 49.52 & 94.28\% & 98.04\% & 99.22\% \\
7: L Elbow & \textbf{48.61} & 89.15\% & 93.61\% & 95.70\% & 50.48 & 89.26\% & 94.44\% & 96.57\% & 54.23 & 89.88\% & 94.92\% & 97.03\% & 51.56 & 91.08\% & 95.14\% & 96.92\% \\
8: R Wrist & \textbf{58.02} & 93.94\% & 96.84\% & 97.85\% & 60.27 & 92.94\% & 96.14\% & 97.54\% & 65.99 & 92.88\% & 95.87\% & 97.24\% & 61.10 & 92.55\% & 96.06\% & 97.50\% \\
9: L Wrist & \textbf{63.14} & 92.49\% & 96.46\% & 98.06\% & 65.66 & 93.79\% & 97.31\% & 98.56\% & 67.68 & 92.16\% & 96.81\% & 98.52\% & 65.51 & 94.21\% & 97.31\% & 98.36\% \\
10: T1 & \textbf{26.09} & 89.07\% & 94.70\% & 96.77\% & 28.89 & 87.00\% & 93.53\% & 96.02\% & 32.31 & 87.87\% & 93.57\% & 95.98\% & 31.01 & 88.93\% & 94.55\% & 96.46\% \\
11: Carina & \textbf{34.49} & 83.06\% & 89.57\% & 92.33\% & 36.81 & 81.79\% & 88.55\% & 91.49\% & 39.53 & 81.90\% & 88.36\% & 91.37\% & 38.56 & 81.67\% & 88.22\% & 91.50\% \\
12: R Hemidiaphragm & \textbf{33.39} & 92.01\% & 95.84\% & 97.34\% & 34.89 & 91.77\% & 95.78\% & 97.37\% & 38.24 & 90.09\% & 95.25\% & 97.37\% & 37.29 & 90.13\% & 94.56\% & 96.45\% \\
13: L Hemidiaphragm & \textbf{34.42} & 90.99\% & 95.97\% & 97.81\% & 36.44 & 91.33\% & 96.09\% & 97.82\% & 39.04 & 90.78\% & 95.96\% & 97.90\% & 38.89 & 89.38\% & 94.55\% & 96.60\% \\
14: T12 & \textbf{32.78} & 92.79\% & 96.85\% & 98.16\% & 35.73 & 92.24\% & 96.54\% & 98.16\% & 36.39 & 91.64\% & 96.44\% & 98.02\% & 36.42 & 90.55\% & 95.46\% & 97.50\% \\ \bottomrule
\end{tabular}}
\vspace{-5mm}
\end{table*}
\subsection{Implementation Detail}
All models were implemented using PyTorch~\cite{pytorch} and trained on a single NVIDIA V100 GPU.  We used a batch size of 128 and trained each model for 50 epochs, where the loss converged to a stable value. Input X-ray images were resized to $224\times 224$ pixels, and optimization was performed using the AdamW optimizer~\cite{adamw}. 

For conformal prediction, we evaluated three significance levels $\alpha \in \{0.1, 0.05, 0.03\}$, corresponding to prediction confidence levels of 90\%, 95\%, and 97\%, respectively. These values determine the desired coverage probability for the prediction regions generated during inference. For Monte Carlo Dropout, we perform $T=20$ stochastic forward passes per sample to estimate the epistemic uncertainty, and the dropout probability was set to $p=0.3$. Unless otherwise noted, all reported results are computed on the test set described in~\cref{sec:dataset}, with calibration-specific metrics evaluated on the calibration set.

\subsection{Evaluation Metrics}
\noindent\textbf{Mean Euclidean Distance: }
For a quantitative evaluation of the C-arm positioning, we use the Euclidean distance between the predicted and ground-truth landmark locations. This measures how far the predicted C-arm displacement places the system from the true landmark. We compute the mean distance across all landmarks and test samples to get a comprehensive evaluation. The measure is defined as follows,
\begin{equation}
    \label{eq:euclidean}
    \begin{aligned}    
    \text{mean}(d_k^{(i)}) = \frac{1}{n \cdot m} \sum_{i=1}^{n} \sum_{k=1}^{m} d(\mathbf{P}_k^{(i)},\hat{\mathbf{P}}_k^{(i)})
\end{aligned}
\end{equation}
Where $n$ is the number of test samples, $m$ is the number of anatomical landmarks. These distances are calculated between the ground-truth position of the landmarks $\mathbf{P}$ and the corresponding predicted positions after applying the model's estimated displacements $\hat{\mathbf{P}}$. For more fine-grained analysis, we also report per-landmark mean distances by omitting the inner average across $m$ landmarks. Ideally, this metric should approach zero, indicating perfect alignment between predicted and actual C-arm positions.

\noindent\textbf{Negative-log Likelihood: }
As we trained probabilistic models that represent a Gaussian distribution over displacements (\cref{eq:nll}), it is natural to include the NLL loss for evaluation. This loss not only quantifies the deviation of the predicted means from the ground truth targets but also assesses the quality of the predicted variance, as it penalizes overconfident predictions. While the NLL may lack an intuitive interpretation compared to spatial errors, it provides a reliable quantitative measure of the overall model fit. For comparisons, lower NLL values indicate better predictive performance and more reliable variance estimates.

\noindent\textbf{Prediction Region Coverage Probability (PRCP): }
We evaluate the reliability of the prediction regions by defining the Prediction Region Coverage Probability (PRCP) metric. PRCP quantifies the proportion of test-time predictions where the ground-truth landmark lies within the calibrated error sphere defined around the model’s output (see Eq.~\ref{eq:region}). Since these radii are computed from empirical quantiles of calibration errors, PRCP directly reflects how well the learned error bounds generalize to unseen data.

As the number of test samples increases, the empirical PRCP should converge to the target coverage level $1-\alpha$. For instance, with $\alpha=0.05$, the ideal PRCP would approach 95\%. Formally, the PRCP is given as follows,
\begin{equation}
    \label{eq:PRCP}
    PRCP = \frac{1}{n \cdot m}\sum_{i=1}^n \sum_{k=1}^m \ \mathbb{I} \{d(\Delta\mathbf{P}_k^{(i)}, \Delta\hat{\mathbf{P}}_k^{(i)}) \in R_k^{(i)}\}
\end{equation}
Where $R_k^{(i)}$ is the spherical region from~\cref{eq:region} that depends on the chosen 1-$\alpha$ probability, and $\mathbb{I}$ is the indicator function. For more detailed analysis, per-landmark PRCP can also be computed.
\begin{table}[!b]
\vspace{-5mm}
\centering
\caption{All augmentation strengths improved ResNet101 performance compared to not applying it.}
\label{tab:aug}
\resizebox{\columnwidth}{!}{%
\begin{tabular}{@{}ccccccc@{}}
\toprule
\multirow{2}{*}{\begin{tabular}[c]{@{}c@{}}Augmentation\\ Strength\end{tabular}} &
  \multirow{2}{*}{\begin{tabular}[c]{@{}c@{}}Mean distance\\from GT (mm) $\downarrow$\end{tabular}} &
  \multicolumn{2}{c}{NLL $\downarrow$} &
  \multicolumn{3}{c}{PRCP} \\ \cmidrule(lr){3-4} \cmidrule(lr){5-7} 
       &                & Calibration    & Test           & 90\%    & 95\%    & 97\%   \\ \midrule
none   & 42.28          & -2.14          & -2.18          & 89.86\% & 95.02\% & 96.99\% \\
weak   & 40.91          & -2.18          & -2.21          & 89.74\% & 94.71\% & 96.69\% \\
mid    & \textbf{39.14} & \textbf{-2.22} & \textbf{-2.28} & 90.04\% & 94.79\% & 96.73\% \\
strong & 41.04          & -2.17          & -2.20          & 89.33\% & 94.57\% & 96.64\% \\ \bottomrule
\end{tabular}}
\vspace{-5mm}
\end{table}

\subsection{Quantitative results}
\noindent\textbf{Backbone Comparison:} To give a comprehensive analysis of the proposed pipeline, we compare multiple different backbones. We test ResNet34 and ResNet101~\cite{resnet} (standard backbones in medical imaging), ConvNeXt Base~\cite{convnext} (a modern convolutional architecture), and ViT Base/16~\cite{vit} (transformer-based model). For comparisons, we rely on the Euclidean distance, NLL, and PRCP. These metrics show both positioning accuracy and confidence modeling. 

~\cref{tab:backbones} presents the quantitative results. ConvNeXt Base achieves the best performance, with a mean distance of 38.6 mm and NLL of -2.18 (calibration) and -2.27 (test) across all landmarks. Other benchmarks show comparative performance with differences not exceeding 6 mm. All models offer impressive PRCP values that closely match the calibrated confidence levels $1-\alpha$. This signifies the generality of our pipeline regardless of the chosen backbone network.

\noindent\textbf{Per landmark Performance:} For a more detailed analysis, we report per-landmark results. These results are relevant in practice, as different use cases may prioritize different regions. From ~\cref{tab:landmarks}, the performance varies across landmarks, where moving ones are generally more challenging to track, e.g., wrists and elbows. However, the PRCP values are consistent across all landmarks, except the Carina. We hypothesize this is due to overconfident bounds in that region. We leave a more detailed analysis of this phenomenon for future work.
\begin{figure*}[!h]
    \centering
    \includegraphics[width=\linewidth]{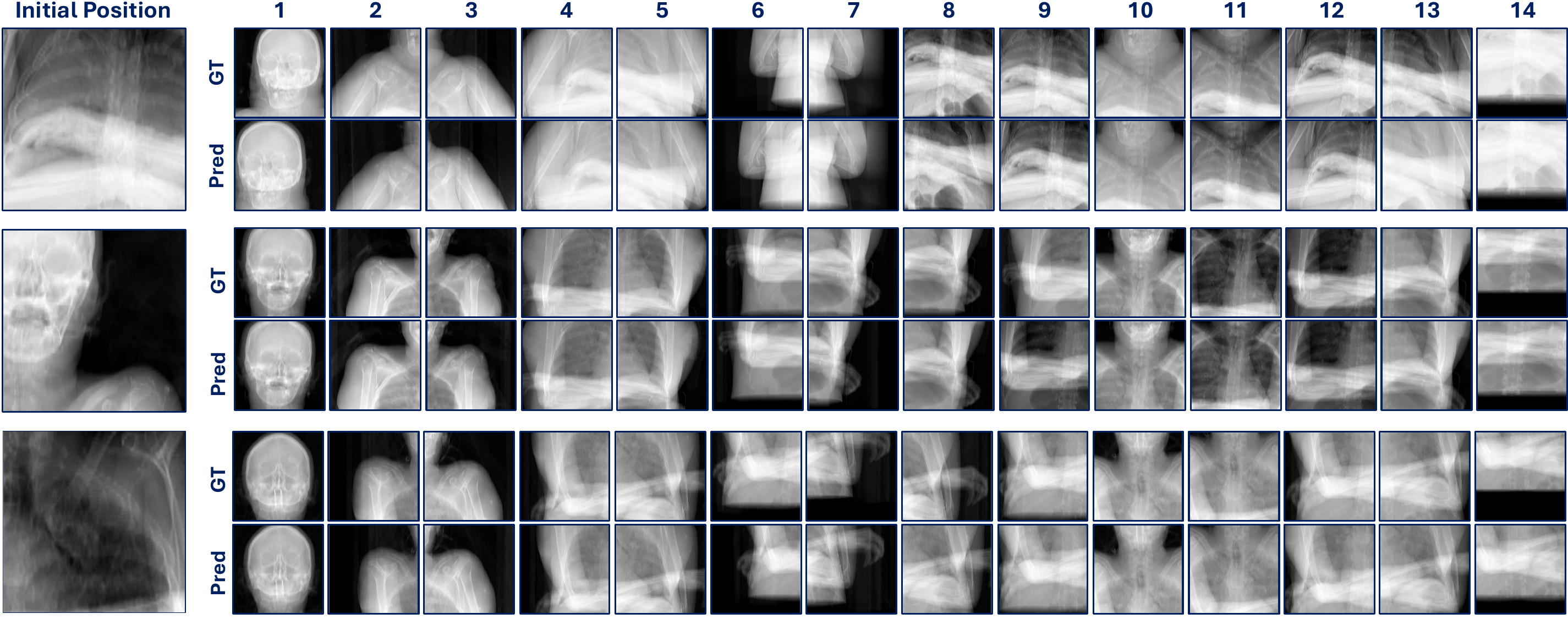}
    \caption{C-arm movement visualization. In all landmarks, the model was successful in localizing the C-arm onto the corresponding ROI.}
    \label{fig:predictions}
    \vspace{-5mm}
\end{figure*}

\noindent\textbf{Skeleton Pose Loss:}
We introduced the skeleton pose loss in~\cref{eq:loss} to regularize the model, encouraging anatomically plausible landmark predictions. To evaluate the effect of this regularization, we vary the loss weight $\lambda$, which controls the influence of the skeleton term in the total loss function. \Cref{tab:skeleton_loss} presents results across different $\lambda$ values, from $\lambda=0$ (no regularization) to $\lambda=10$ (strong regularization). As $\lambda$ increases, model performance improves consistently across all metrics, which shows that utilizing anatomical priors enhances both localization and confidence accuracy.
\begin{table}[!b]
\centering
\caption{Changing the weight of the skeleton pose loss with ResNet101. As $\lambda$ increases, the performance improves across all metrics.}
\label{tab:skeleton_loss}
\resizebox{\columnwidth}{!}{%
\begin{tabular}{@{}ccccccc@{}}
\toprule
\multirow{2}{*}{$\lambda$} &
  \multirow{2}{*}{\begin{tabular}[c]{@{}c@{}}Mean distance \\ from GT (mm) $\downarrow$\end{tabular}} &
  \multicolumn{2}{c}{NLL $\downarrow$} &
  \multicolumn{3}{c}{PRCP} \\ \cmidrule(lr){3-4} \cmidrule(lr){5-7} 
    &                & Calibration    & Test           & 90\%    & 95\%    & 97\%   \\ \midrule
0   & 41.47          & -2.15          & -2.20          & 89.67\% & 94.85\% & 96.88\% \\
0.5 & 42.05          & -1.99          & -2.02          & 89.60\% & 94.99\% & 96.97\% \\
1   & 40.91          & -2.18          & -2.21          & 89.74\% & 94.71\% & 96.69\% \\
5   & 40.23          & -2.23          & -2.27          & 89.24\% & 94.46\% & 96.56\% \\
10  & \textbf{39.28} & \textbf{-2.28} & \textbf{-2.35} & 90.20\% & 95.04\% & 96.94\% \\ \bottomrule
\end{tabular}}
\vspace{-5mm}
\end{table}

\noindent\textbf{Data Augmentation:}
As described in~\cref{sec:methods}, we introduced a data augmentation scheme that randomly positions the patient on the operating table. This mimics a real clinical setting where the initial position of the patient is not constant. We compare three levels of augmentation: weak, medium, and strong, where each gradually increases the magnitude of a patient's shift across the table. ~\cref{tab:aug} shows our comparison, where we can see that including data augmentation, regardless of its strength, improves performance. The medium level is the empirically optimal one.

\subsection{Qualitative Results} 
Qualitative results of our pipeline are presented in~\cref{fig:predictions}. The left images show randomly selected initial test X-rays, while the right images visualize the simulated C-arm position after applying the predicted displacements. The model successfully aligns the C-arm above all ROIs, demonstrating its localization ability in a simulated environment.
\subsection{Application: Stroke Thrombectomy}
To demonstrate a practical application of our C-arm pipeline, we design an experiment to explore the following question: \textit{Does taking multiple steps improve localization accuracy?} We simulate a clinical scenario involving stroke thrombectomy, where the ROI is the skull (Landmark 1), and intermediate steps involve targeting the T1 (10) and the Carina (11). These landmarks were chosen due to their rigid anatomical alignment, making them relatively fixed with respect to the skull. 
We design a multi-step prediction pipeline, where the output of each stage serves as the input for the subsequent stage, simulating iterative repositioning toward the target. In this experiment, we compare four pre-defined paths: direct prediction to landmark \textbf{1}, and three multi-stage paths:\textbf{10$\rightarrow$1}, \textbf{11$\rightarrow$1}, and \textbf{11$\rightarrow$10$\rightarrow$1}.
\begin{table}[t]
\centering
\caption{MAE and variance of 2D absolute error in the stroke thrombectomy application.}
\label{tab:exp_landmark}
\resizebox{\columnwidth}{!}{%
\begin{tabular}{@{}lcccc@{}}
\toprule
\multirow{2}{*}{Metric} & \multicolumn{4}{c}{Landmark Path} \\
\cmidrule(lr){2-5}
 & 1 & 10$\rightarrow$1 & 11$\rightarrow$1 & 11$\rightarrow$10$\rightarrow$1 \\
\midrule
MAE $\downarrow$ & 20.81 & \textbf{18.97} & 21.30 & 19.27 \\
Absolute Error Variance & 211.24 & 130.89 & 144.34 & 137.32 \\
\bottomrule
\end{tabular}}
\vspace{-5mm}
\end{table}

\begin{figure}[!htbp]
    \centering
    \includegraphics[width=\linewidth]{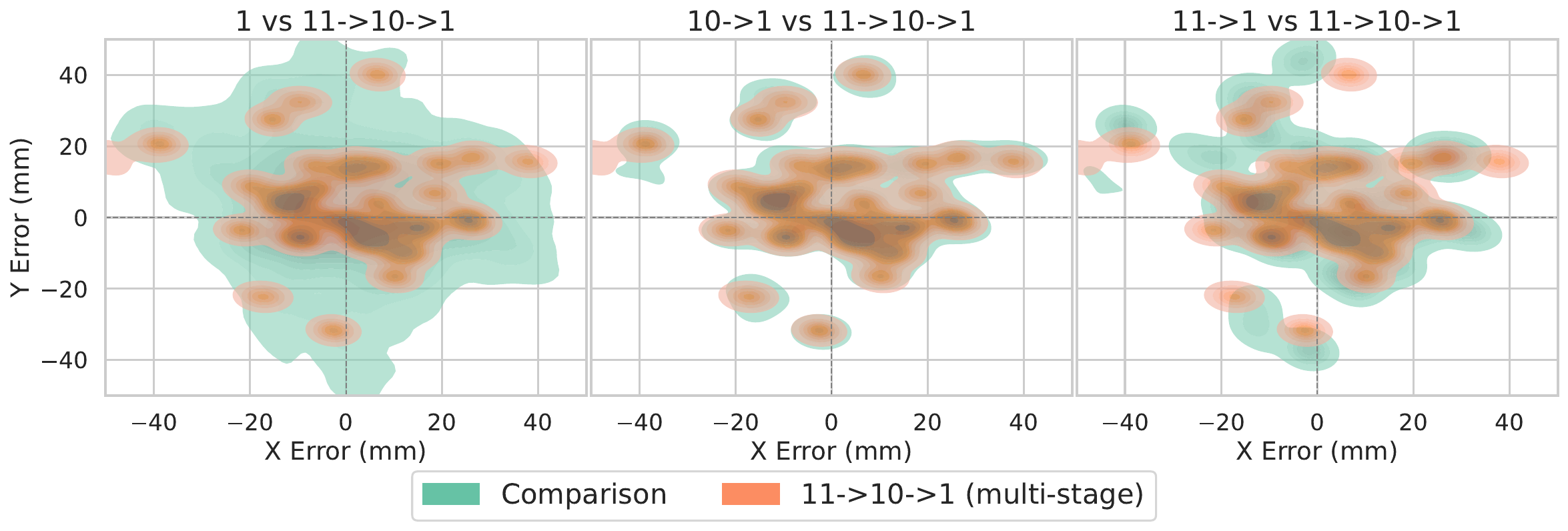}
    \caption{Kernel Density Estimates (KDE) of prediction errors in $x$ and $y$ directions. Long tails, due to outliers, were clipped for visibility.}
    \label{fig:kdel}
    \vspace{-5mm}
\end{figure}

As shown in Table~\cref{tab:exp_landmark}, utilizing intermediate landmark predictions reduces the 2D Mean Absolute Error (MAE) compared to direct prediction. It also influences the convergence quality, which is seen in the error variance values. A single-shot prediction to the skull leads to a much higher error variance value than all other multi-step predictions, meaning that the latter approach is more stable.~\cref{fig:kdel} illustrates the test error distribution across the vertical and horizontal axes. 
Compared to single-shot prediction, the multi-stage path 11$\rightarrow$10$\rightarrow$1 achieves better error convergence, closer to 0. Further investigation is needed to explore different intermediate landmark combinations to reach different ROIs, which we leave for future work.

\section{Conclusion and Future Work}
\label{sec:conclusion}
This work proposed a pipeline for C-arm control to achieve safe and reliable automation in interventional fluoroscopy. We see this contribution as a step toward fully autonomous C-arm systems capable of moving from any initial point to any desired target. We focused on confidence and uncertainty quantification, which we view as potential sub-modules that can be integrated into broader autonomous systems. Our method demonstrated strong localization accuracy and reliable confidence calibration across a range of backbone architectures. However, as shown in ~\cref{tab:landmarks}, the errors vary across landmarks, with larger errors observed for anatomically dynamic regions. This suggests that a deeper analysis of landmark-specific variability is needed for clinical deployment, which we leave for future work.

Although the C-arm operates with 6 DoF, our current implementation controls only 3 translational axes. Extending the framework to include rotational control will be a key area of future research. Moreover, validating the applicability of our approach across various interventional procedures beyond Stroke Thrombectomy would enhance its generalizability. We plan to investigate broader procedures that benefit from uncertainty-aware automated C-arm positioning. 

Another important step is to validate the framework with real X-ray data in place of synthetic DRRs. Real X-rays contain noise and artifacts that are absent in synthetic data, which better reflect operating settings. Finally, transferring the learned policies from simulation to a physical C-arm will be a natural extension, requiring engineering challenges and hardware interfacing.
\section{Acknowledgments}
\label{sec:acknowledgments}
\vspace{-0.25cm}
This work was supported by the National Science Foundation under Grants No. 2218063.
\vspace{-4pt}

{
    \small
    \bibliographystyle{ieeenat_fullname}
    \bibliography{ref}
}

\end{document}